\documentclass[runningheads]{llncs}

\usepackage{graphicx}
\usepackage{amsmath}
\usepackage{amssymb}
\usepackage{mathtools}
\usepackage{hyperref}
\usepackage[ruled,vlined]{algorithm2e}
\usepackage{cite}
\usepackage{paralist}
\usepackage{color}
\usepackage{mwe}
\usepackage{graphbox}
\usepackage{stmaryrd}
\usepackage[lighttt]{lmodern}
\usepackage{tabularx}

\newcommand{\cond}[1]{\text{``}#1\text{''}}
\newcommand{\ncond}[1]{\overline{\cond{#1}}}
\newcommand{\tp}{\textit{TP}}
\newcommand{\fp}{\textit{FP}}
\newcommand{\tn}{\textit{TN}}
\newcommand{\fn}{\textit{FN}}

\begin{document}

\title{On the Trade-off Between Consistency and Coverage in Multi-label Rule Learning Heuristics}
\titlerunning{Consistency and Coverage in Multi-label Rule Learning Heuristics}

\author{
\textit{\textbf{Preprint version}. To appear in Proceedings of the 22nd International Conference on Discovery Science, 2019}\\ \ \\ 
Michael Rapp \and Eneldo Loza Menc\'ia \and Johannes F\"urnkranz}
\authorrunning{M. Rapp \and E. Loza Menc\'ia \and J. F\"urnkranz}

\institute{Knowledge Engineering Group, TU Darmstadt, Germany\\ \email{\{mrapp, eneldo, juffi\}@ke.tu-darmstadt.de}}

\maketitle

\begin{abstract}

Recently, several authors have advocated the use of rule learning algorithms to model multi-label data, as rules are interpretable and can be comprehended, analyzed, or qualitatively evaluated by domain experts. Many rule learning algorithms employ a heuristic-guided search for rules that model regularities contained in the training data and it is commonly accepted that the choice of the heuristic has a significant impact on the predictive performance of the learner. Whereas the properties of rule learning heuristics have been studied in the realm of single-label classification, there is no such work taking into account the particularities of multi-label classification. This is surprising, as the quality of multi-label predictions is usually assessed in terms of a variety of different, potentially competing, performance measures that cannot all be optimized by a single learner at the same time. In this work, we show empirically that it is crucial to trade off the consistency and coverage of rules differently, depending on which multi-label measure should be optimized by a model. Based on these findings, we emphasize the need for configurable learners that can flexibly use different heuristics. As our experiments reveal, the choice of the heuristic is not straight-forward, because a search for rules that optimize a measure locally does usually not result in a model that maximizes that measure globally.

\end{abstract}

\section{Introduction}
\label{sec_introduction}

As many real-world classification problems require to assign more than one label to an instance, multi-label classification (MLC) has become a well-established topic in the machine learning community. There are various applications of MLC such as text categorization \cite{lewis1992, klimt2004}, the annotation of images \cite{boutell2004, li2008} and music \cite{trohidis2008, turnbull2008}, as well as use cases in bioinformatics \cite{diplaris2005} and medicine \cite{pestian2007}.

Rule learning algorithms are a well-researched approach to solve classification problems \cite{furnkranz2012}. In comparison to complex statistical methods, like for example support vector machines or artificial neural networks, their main advantage is the interpretability of the resulting models. Rule-based models can easily be understood by humans and form a structured hypothesis space that can be analyzed and modified by domain experts. Ideally, rule-based approaches are able to yield insight into the application domain by revealing patterns and regularities hidden in the data and allow to reason why individual predictions have been made by a system. This is especially relevant in safety-critical domains, such as medicine, power systems, or financial markets, where malfunctions and unexpected behavior may entail the risk of health damage or financial harm.

\subsubsection{Motivation and goals.}
\label{sec_motivation}

To assess the quality of multi-label predictions in terms of a single score, several commonly used performance measures exist. Even though some of them originate from measures used in binary or multi-class classification, different ways to aggregate and average the predictions for individual labels and instances --- most prominently \emph{micro-} and \emph{macro-averaging} --- exist in MLC. Some measures like \emph{subset accuracy} are even unique to the multi-label setting. No studies that investigate the effects of using different rule learning heuristics in MLC and discuss how they affect different multi-label performance measures have been published so far.

In accordance with previous publications in single-label classification, we argue that all common rule learning heuristics basically trade off between two aspects, \emph{consistency} and \emph{coverage} \cite{furnkranz2005}. Our long-term goal is to better understand how these two aspects should be weighed to assess the quality of candidate rules during training if one is interested in a model that optimizes a certain multi-label performance measure. As a first step towards this goal, we present a method for flexibly creating rule-based models that are built with respect to certain heuristics. Using this method, we empirically analyze how different heuristics affect the models in terms of predictive performance and model characteristics. We demonstrate how models that aim to optimize a given multi-label performance measure can deliberately be trained by choosing a suitable heuristic. By comparing our results to a state-of-the-art rule learner, we emphasize the need for configurable approaches that can flexibly be tailored to different multi-label measures. Due to space limitations, we restrict ourselves to micro-averaged measures, as well as to Hamming and subset accuracy.

\subsubsection{Structure of this work.}
\label{sec_structure}

We start in Section~\ref{sec_preliminaries} by giving a formal definition of multi-label classification tasks as well as an overview of inductive rule learning and the rule evaluation measures that are relevant to this work. Based on these foundations, in Section \ref{sec_algorithm}, we discuss our approach for flexibly creating rule-based classifiers that are built with respect to said measures. In Section~\ref{sec_evaluation}, we present the results of the empirical study we have conducted, before we provide an overview of related work in Section~\ref{sec_related_work}. Finally, we conclude in Section~\ref{sec_conclusion} by recapitulating our results and giving an outlook on planned future work.

\section{Preliminaries}
\label{sec_preliminaries}

MLC is a supervised learning problem in which the task is to associate an instance with one or several labels $\lambda_i$ out of a finite label space $\mathbb{L} = \left \{ \lambda_1, \dots, \lambda_n \right \}$, with $n = \left| \mathbb{L} \right|$ being the total number of predefined labels. An individual instance $\boldsymbol{x}_j$ is represented in attribute-value form, i.e., it consists of a vector $\boldsymbol{x}_j = \left( v_1, \dots, v_l \right) \in \mathbb{D} = A_1 \times \dots \times A_l$, where $A_i$ is a numeric or nominal attribute. Additionally, each instance $\boldsymbol{x}_j$ is associated with a binary label vector $\boldsymbol{y}_j = \left( y_1, \dots, y_n \right) = \left \{ 0, 1 \right \}^n$, where $y_i$ indicates the presence ($1$) or absence ($0$) of label $\lambda_i$.
Consequently, the training data set of a MLC problem can be defined as a set of tuples $T = \left\{ \left( \boldsymbol{x}_1, \boldsymbol{y}_1 \right), \dots, \left( \boldsymbol{x}_m, \boldsymbol{y}_m \right) \right\}$, with $m = \left| T \right|$ being the number of available training instances. The classifier function $g \left( . \right)$, that is deduced from a given training data set, maps an instance $\boldsymbol{x}$ to a predicted label vector $\boldsymbol{\hat{y}} = \left( \hat{y}_1, \dots, \hat{y}_n \right) = \left \{ 0, 1 \right \}^n$.

\subsection{Classification rules}
\label{sec_rules}

In this work, we are concerned with the induction of conjunctive, propositional rules $\boldsymbol{r}: H \leftarrow B$. The body $B$ of such a rule consists of one or several conditions that compare an attribute-value $v_i$ of an instance to a constant by using a relational operator such as $=$ (in case of nominal attributes), or $<$ and $\geq$ (in case of numerical attributes). On the one hand, the body of a conjunctive rule can be viewed as a predicate $B: \boldsymbol{x} \rightarrow \left \{ \text{true}, \text{false} \right \}$ that states whether an instance $\boldsymbol{x}$ satisfies all of the given conditions, i.e., whether the instance is \emph{covered} by the rule or not. On the other hand, the head $H$ of a (single-label head) rule consists of a single label assignment ($\hat{y}_i = 0$ or $\hat{y}_i = 1$) that specifies whether the label $\lambda_i$ should be predicted as present ($1$) or absent ($0$).

\subsection{Binary relevance method}
\label{sec_binary_relevance}

In the present work, we use the \emph{binary relevance} transformation method (cf.~\cite{boutell2004}), which reduces MLC to binary classification by treating each label $\lambda_i \in \mathbb{L}$ of a MLC problem independently. For each label $\lambda_i$, we aim at learning rules that predict the minority class $t_i \in \left \{ 0, 1 \right \}$, i.e., rules that contain the label assignment $\hat{y}_i = t_i$ in their head. We define $t_i = 1$, if the corresponding label $\lambda_i$ is associated with less than 50\% of the training instances, or $t_i = 0$ otherwise.

A rule-based classifier --- also referred to as a \emph{theory} --- combines several rules into a single model. In this work, we use (unordered) rule sets containing all rules that have been induced for the individual labels. Such a rule set can be considered as a disjunction of conjunctive rules (DNF). At prediction time, all rules that cover a given instance are taken into account to determine the predicted label vector $\boldsymbol{\hat{y}}$. An individual element $\hat{y}_i \in \boldsymbol{\hat{y}}$, that corresponds to the label $\lambda_i$, is set to the minority class $t_i$ if at least one of the covering rules contains the label assignment $\hat{y}_i = t_i$ in its head. Otherwise, the element is set to the majority class $1 - t_i$. As all rules that have been induced for a label $\lambda_i$ have the same head, no conflicts may arise in the process.

\subsection{Bipartition evaluation functions}
\label{sec_bipartitions}

To assess the quality of individual rules, usually bipartition evaluation functions $\delta: \mathbb{N}^{2 \times 2} \rightarrow \mathbb{R}$ are used \cite{tsoumakas2009}. Such functions --- also called \emph{heuristics} --- map a two-dimensional confusion matrix to a heuristic value $h \in \left[ 0, 1 \right]$. A confusion matrix consists of the number of \emph{true positive} ($\tp$), \emph{false positive} ($\fp$), \emph{true negative} ($\tn$), and \emph{false negative} ($\fn$) labels that are predicted by a rule. We calculate the example-wise aggregated confusion matrix $C_{\boldsymbol{r}}$ for a rule $\boldsymbol{r}: \hat{y}_i \leftarrow B$ as
\begin{equation}
\label{eq_confusion_matrix}
\begin{split}
C_{\boldsymbol{r}} & \coloneqq \left( \begin{array}{cc}
  \tp & \fp \\
  \fn & \tn
\end{array} \right) = C_i^1 \oplus \dots \oplus C_i^j \oplus \dots \oplus C_i^m
\end{split}
\end{equation}
where $\oplus$ denotes the cell-wise addition of atomic confusion matrices $C_i^j$ that correspond to label $\lambda_i$ and instance $\boldsymbol{x}_j$.

Further, let $y_i^j$ and $\hat{y}_i^j$ denote the absence ($0$) or presence ($1$) of label $\lambda_i$ for an instance $\boldsymbol{y}_j$ according to the ground truth and a rule's prediction, respectively. Based on these variables, we calculate the elements of $C_i^j$ as
\begin{equation}
\label{eq_atomic_confusion_matrix}
    \begin{array}{cc}
        \tp_i^j = \llbracket y_i^j = t_i \wedge \hat{y}_i^j = t_i \rrbracket \quad & \fp_i^j = \llbracket y_i^j \neq t_i \wedge \hat{y}_i^j = t_i \rrbracket \\
        \fn_i^j = \llbracket y_i^j = t_i \wedge \hat{y}_i^j \neq t_i \rrbracket \quad & \tn_i^j = \llbracket y_i^j \neq t_i \wedge \hat{y}_i^j \neq t_i \rrbracket
    \end{array}
\end{equation}
where $\llbracket x \rrbracket = 1$, if $x$ is true, $0$ otherwise.

\subsection{Rule learning heuristics}
\label{sec_heuristics}

A good rule learning heuristic should (among other aspects) take both, the \emph{consistency} and \emph{coverage} of a rule, into account \cite{janssen2010, furnkranz2012}. On the one hand, rules should be consistent, i.e., their prediction should be correct for as many of the covered instances as possible. On the other hand, rules with great coverage, i.e., rules that cover a large number of instances, tend to be more reliable, even though they may be less consistent.

The \emph{precision} metric exclusively focuses on the consistency of a rule. It calculates as the fraction of correct predictions among all covered instances:
\begin{equation}
\label{eq_precision}
\begin{split}
\delta_{prec} \left( C \right) \coloneqq & \frac{\tp}{\tp + \fp}
\end{split}
\end{equation}

In contrast, \emph{recall} focuses on the coverage of a rule. It measures the fraction of covered instances among all --- covered and uncovered --- instances for which the label assignment in the rule's head is correct:
\begin{equation}
\label{eq_recall}
\begin{split}
\delta_{rec} \left( C \right) \coloneqq & \frac{\tp}{\tp + \fn}
\end{split}
\end{equation}

The \emph{F-measure} calculates as the (weighted) harmonic mean of precision and recall. It allows to trade off the consistency and coverage of a rule depending on the user-configurable parameter $\beta$:
\begin{equation}
\label{eq_fmeasure}
\begin{split}
\delta_F \left( C \right) \coloneqq & \frac{\beta^2 + 1}{\frac{\beta^2}{\delta_{rec} \left( C \right)} + \frac{1}{\delta_{prec} \left( C \right)}} \text{, with } \beta \in \left[ 0, +\infty \right]
\end{split}
\end{equation}

As an alternative to the F-measure, we use different parameterizations of the \emph{m-estimate} in this work. It is defined as
\begin{equation}
\label{eq_mestimate}
\begin{split}
\delta_m \left( C \right) \coloneqq & \frac{\tp + m \cdot \frac{P}{P + N}}{\tp + \fp + m} \text{, with } m \geq 0
\end{split}
\end{equation}
where $P = \tp + \fn$ and $N = \fp + \tn$. Depending on the parameter $m$, this measure trades off precision and \emph{weighted relative accuracy} (WRA). If $m = 0$, it is equivalent to precision and therefore focuses on consistency. As $m$ approaches $+\infty$, it converges to WRA and puts more emphasis on coverage, respectively \cite{furnkranz2012}.

\section{Induction of rule-based theories}
\label{sec_algorithm}

For our experimental study, we implemented a method that allows to generate a large number of rules for a given training data set in a short amount of time (cf.~Section~\ref{sec_rule_generation}).\footnote{Source code available at \url{https://github.com/mrapp-ke/RuleGeneration}.} The rules should ideally be unbiased, i.e., they should not be biased in favor of a certain heuristic, and they should be diverse, i.e., general rules should be included as well as specific rules. Given that these requirements are met, we consider the generated rules to be representative samples for the space of all possible rules, which is way too large to be explored exhaustively. We use the generated candidate rules as a starting point for building different theories. They consist of a subset of rules that are selected with respect to a specific heuristic (cf.~Section~\ref{sec_candidate_selection}) and filtered according to a threshold (cf.~Section~\ref{sec_thresholding}). Whereas the first step yields a theory with great coverage, the threshold selection aims at improving its consistency.

\subsection{Generation of candidate rules}
\label{sec_rule_generation}

As noted in Section~\ref{sec_binary_relevance}, we consider each label $\lambda_i \in \mathbb{L}$ of a MLC problem independently. For each of the labels we train multiple random forests \cite{breiman2001}, using varying configuration parameters, and extract rules from their decision trees.\footnote{We use the random forest implementation provided by Weka 3.9.3, which is available at \url{https://www.cs.waikato.ac.nz/ml/weka}.} As illustrated in Algorithm~\ref{alg_rule_generation}, we repeat the process until a predefined number of rules $\gamma$ has been generated. 

Each random forest consists of a predefined number of decision trees (we specify $I = 10$). To ensure that we are able to generate diverse rules later on, we vary the configuration parameter $depth \in \left[ 0, 8 \right]$ that specifies the maximum depth of trees (unrestricted, if $depth = 0$) (cf. Algorithm~\ref{alg_rule_generation}, \texttt{trainForest}). For building individual trees, we only take a subset of the available training instances and attributes into account, which guarantees a diverse set of trees. Bagging is used for sampling the training instances, i.e., if $m$ instances are available in total, $m \cdot P$ instances ($P = 100\%$, by default) are drawn randomly with replacement. Additionally, each time a new node is added to a decision tree, only a random selection of $K$ out of $l$ attributes ($K = \log_2 \left( l - 1 \right) + 1$, by default) is considered.

\begin{algorithm}[t]
\label{alg_rule_generation}
\SetKwFunction{TrainForest}{trainForest}\SetKwFunction{ExtractRules}{extractRules}
\SetKwInOut{Input}{input}\SetKwInOut{Output}{output}
  \Input{min. number of rules to be generated $\gamma$}
  \Output{rule set $R$}
  $R = \emptyset$\\
  \While{$\left|R\right| < \gamma$} {
    \ForEach{$\lambda_i \in \mathbb{L}$ \upshape\textbf{and} $\textit{depth} \in \left[0, 8 \right]$} {
      $\textit{rf}$ = \TrainForest{$\lambda_i$, $\textit{depth}$}\\
      $R$ = $R$ $\cup$ \ExtractRules{$\textit{rf}$}\\
    }
  }
  \Return $R$
  \caption{Iterative generation of rules from random forests}
\end{algorithm}

To extract rules from a random forest (cf.~Algorithm~\ref{alg_rule_generation}, \texttt{extractRules}), we traverse all paths from the root node to a leaf in each of its decision trees. We only consider paths that lead to a leaf where the minority class $t_i$ is predicted. As a consequence, all rules that are generated with respect to a certain label $\lambda_i$ have the same head $\hat{y}_i = t_i$. The body of a rule consists of a conjunction of all conditions encountered on the path from the root to the correspondin

\subsection{Candidate subset selection}
\label{sec_candidate_selection}

Like many traditional rule learning algorithms, we use a \emph{separate-and-conquer} (SeCo) strategy for selecting candidate rules, i.e., new rules are added to the theory until all training instances are covered (or until it describes the training data sufficiently according to some stopping criterion). Whenever a new rule is added to the theory the training instances it covers are removed (``separate'' step), and the next rule is chosen according to its performance on the remaining instances (``conquer'' step).

To create different theories, we select subsets of the rules that have been generated earlier (cf.~Section~\ref{sec_rule_generation}). We therefore apply the SeCo strategy for each label independently, i.e., for each label $\lambda_i$ we take all rules with head $\hat{y}_i = t_i$ into account. Among these candidates we successively select the best rule according to a heuristic $\delta$ (cf.~Section~\ref{sec_heuristics}) until all \emph{positive} training instances $P_i = \left \{ \left( \boldsymbol{x}, \boldsymbol{y} \right) \in T \mid y_i = t_i \right \}$, with respect to label $\lambda_i$, are covered. To measure the quality of a candidate $\boldsymbol{r}$ according to $\delta$, we only take yet uncovered instances into account for computing the confusion matrix $C_{\boldsymbol{r}}$. If two candidates evaluate to the same heuristic value, we prefer the one that
\begin{inparaenum}[a)\upshape]
  \item covers more true positives, or
  \item contains fewer conditions in its body.
\end{inparaenum}
Whenever a new rule is added, the overall coverage of the theory increases, as more positive training instances are covered. However, the rule may also cover some of the \emph{negative} instances $N_i = T \setminus P_i$. As the rule's prediction is incorrect in such cases, the consistency of the theory may decrease.

\subsection{Threshold selection}
\label{sec_thresholding}

As described in Section~\ref{sec_candidate_selection}, we use a SeCo strategy to select more rules until all positive training instances are covered for each label. In this way, the coverage of the resulting theory is maximized at the expense of consistency, because each rule contributes to the overall coverage, but might introduce wrong predictions for some instances. To trade off between these aspects, we allow to (optionally) specify a threshold $\phi$ that aims at diminishing the effects of inconsistent rules. It is compared to a heuristic value that is calculated for each rule according to the heuristic $\delta$. For calculating the heuristic value, the rule's predictions on the entire training data set are taken into account. This is different from the candidate selection discussed in Section~\ref{sec_candidate_selection}, where instances that are already covered by previously selected rules are not considered. Because the candidate selection aims at selecting non-redundant rules, that cover the positive training instances as uniformly as possible, it considers rules in the context of their predecessors. In contrast, the threshold $\phi$ is applied at prediction time when no order is imposed on the rules, i.e., all rules whose heuristic value exceeds the threshold equally contribute to the prediction.

\section{Evaluation}
\label{sec_evaluation}

In this section, we present an empirical study that emphasises the need to use varying heuristics for candidate selection and filtering to learn theories that are tailored to specific multi-label measures. We further compare our method to different baselines to demonstrate the benefits of being able to flexibly adjust a learner to different measures, rather than employing a general-purpose learner.

\subsection{Experimental setup}
\label{sec_experimental_setup}

We applied our method to eight different data sets taken from the Mulan project.\footnote{Data sets and detailed statistics available at \url{http://mulan.sourceforge.net/datasets-mlc.html}.} We set the minimum number of rules to be generated to 300.000 (cf.~Algorithm~\ref{alg_rule_generation}, parameter $\gamma$). For candidate selection according to Section~\ref{sec_candidate_selection}, we used the \mbox{m-estimate} (cf.~Equation~\ref{eq_mestimate}) with $m =  0, 2^1, 2^2, \dots, 2^{19}$. For each of these variants, we applied varying thresholds $\phi$ according to Section~\ref{sec_thresholding}. The thresholds have been chosen such that they are satisfied by at least $100\%, 95\%, \dots, 5\%$ of the selected rules. All results have been obtained using 10-fold cross validation. 

In addition to the m-estimate, we also used the F-measure (cf.~Equation~\ref{eq_fmeasure}) with varying $\beta$-parameters. As the conclusions drawn from these experiments are very similar to those for the m-estimate, we focus on the latter at this point.

Among the performance measures that we report are micro-averaged precision and recall. Given a global confusion matrix $C \coloneqq C_1^1 \oplus \dots \oplus C_i^j \oplus \dots \oplus C_n^m$ that consists of the $\tp$, $\fp$, $\tn$, and $\fn$ aggregated over all test instances $\boldsymbol{x}_j$ and labels $\lambda_i$, these two measures are calculated as defined in Equations~\ref{eq_precision} and~\ref{eq_recall}. Moreover, we report the micro-averaged F1 score (cf.~Equation~\ref{eq_fmeasure} with $\beta = 1$) as well as Hamming and subset accuracy. Hamming accuracy calculates as
\begin{equation}
  \label{eq_hamming_accuracy}
  \begin{split}
    \delta_{Hamm} \left( C \right) & \coloneqq \frac{\tp + \tn}{\tp + \fp + \tn + \fn}
  \end{split}
\end{equation}
whereas subset accuracy differs from the other measures, because it is computed instance-wise. Given true label vectors $Y = \left( \boldsymbol{y}_1, \dots, \boldsymbol{y}_m \right)$ and predicted label vectors $\hat{Y} = \left( \boldsymbol{\hat{y}}_1, \dots, \boldsymbol{\hat{y}}_m \right)$, it measures the fraction of perfectly labeled instances:
\begin{equation}
  \label{eq_subset_accuracy}
  \begin{split}
    \delta_{acc} \left( Y, \hat{Y} \right) & \coloneqq \frac{1}{m} \sum_j \llbracket \boldsymbol{y}_j = \hat{\boldsymbol{y}}_j \rrbracket
  \end{split}
\end{equation}

\subsection{Analysis of different parameter settings}
\label{sec_analysis}

\begin{figure}[b!]
  \centering
  \begin{tabular}{lllll}
    \begin{tabular}[t]{l}
      \multicolumn{1}{c}{$\phi$} \\
      \includegraphics{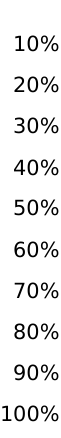} \\
      \\
      \includegraphics{img/y_axis_labels.pdf} \\
    \end{tabular}
    \begin{tabular}[t]{l}
      \multicolumn{1}{c}{\textbf{Avg. ranks ``Hamm. acc.''}} \\
      \includegraphics{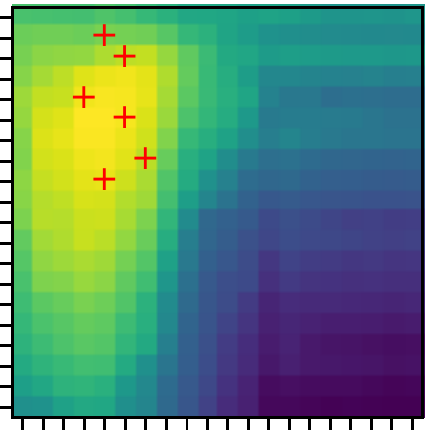} \\ 
      \multicolumn{1}{c}{\textbf{Avg. ranks ``Subset acc.''}} \\
      \includegraphics{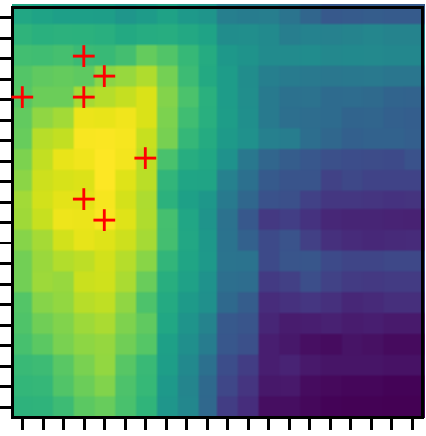} \\ 
      \includegraphics{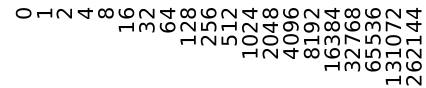} \\
    \end{tabular} &
    \begin{tabular}[t]{l}
      \\
      \includegraphics{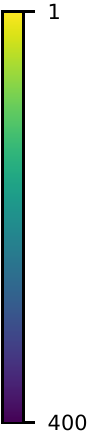} \\ 
      \\
      \includegraphics{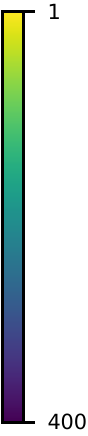} \\
    \end{tabular} &
    \begin{tabular}[t]{l}
      \multicolumn{1}{c}{\textbf{Std.-dev. ``Hamm. acc.''}} \\
      \includegraphics{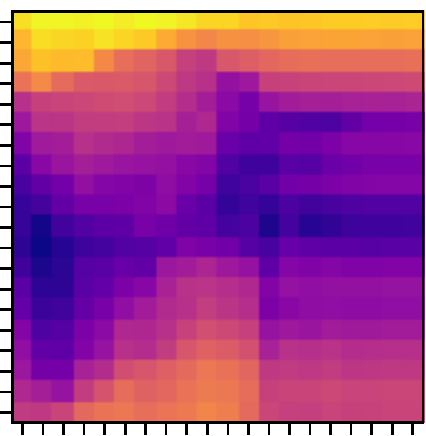} \\
      \multicolumn{1}{c}{\textbf{Std.-dev. ``Subset acc.''}} \\
      \includegraphics{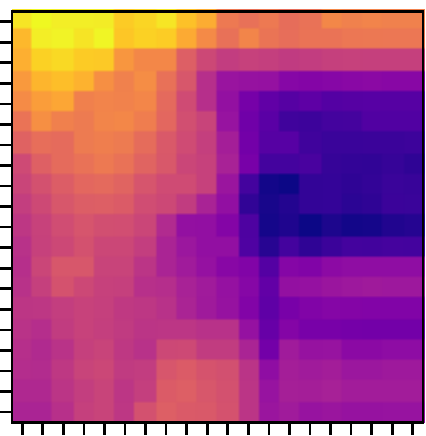} \\
      \includegraphics{img/m-estimate/x_axis_labels.pdf} \\
    \end{tabular} &
    \begin{tabular}[t]{l}
      \\
      \includegraphics{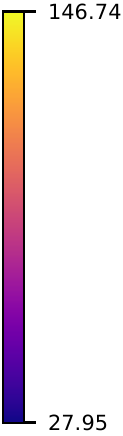} \\
      \\
      \includegraphics{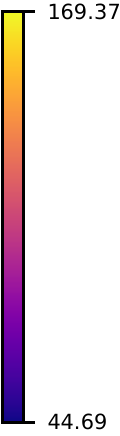} \\
      $m$ \\
    \end{tabular} \\
  \end{tabular}
\caption{Ranks and standard deviation of average ranks over all data sets according to Hamming and subset accuracy using different parameters $m$ (horizontal axis) and $\phi$ (vertical axis). Best parameters for different data sets specified by red \textcolor{red}{+} signs.}
\label{fig_evaluation_avg_hamm_subs}
\end{figure}

For a broad analysis, we trained $20^2=400$ theories per data set using the same candidate rules, but selecting and filtering them differently by using varying combinations of the parameters $m$ and $\phi$ as discussed in Section~\ref{sec_experimental_setup}. We visualize the performance and characteristics of the resulting models as two-dimensional matrices of scores (cf.~e.g.~Figure~\ref{fig_evaluation_avg_hamm_subs}). One dimension corresponds to the used $m$-parameter, the other refers to the threshold $\phi$, respectively.

\begin{figure}[p]
  \centering
  \begin{tabular}{lllll}
    \begin{tabular}[t]{l}
      \multicolumn{1}{c}{$\phi$} \\
      \includegraphics{img/y_axis_labels.pdf} \\
      \\
      \includegraphics{img/y_axis_labels.pdf} \\
      \\
      \includegraphics{img/y_axis_labels.pdf} \\
    \end{tabular}
    \begin{tabular}[t]{l}
      \multicolumn{1}{c}{\textbf{Avg. ranks ``Precision''}} \\
      \includegraphics{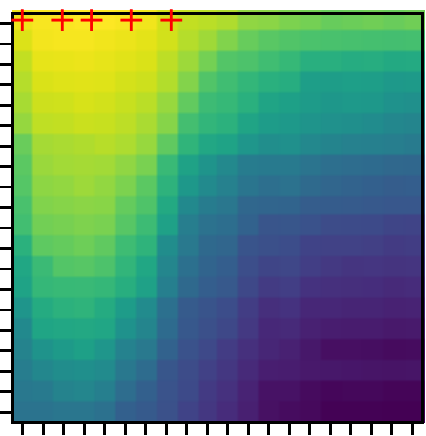} \\
      \multicolumn{1}{c}{\textbf{Avg. ranks ``Recall''}} \\
      \includegraphics{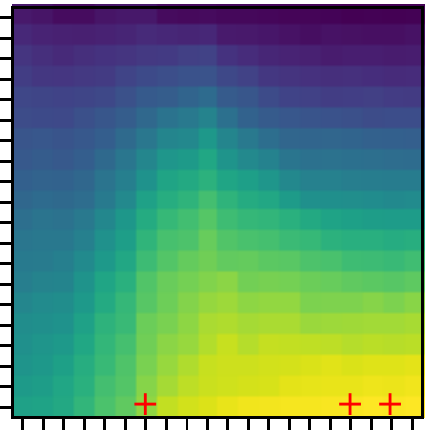} \\
      \multicolumn{1}{c}{\textbf{Avg. ranks ``F1''}} \\
      \includegraphics{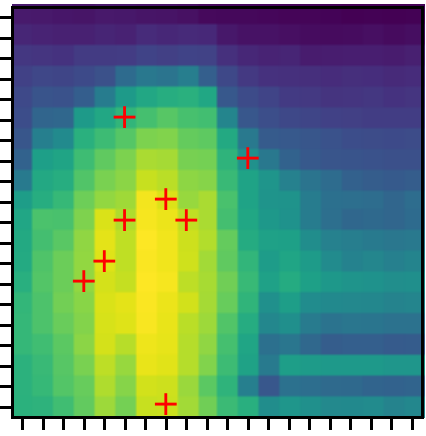} \\
      \includegraphics{img/m-estimate/x_axis_labels.pdf} \\
    \end{tabular} &
    \begin{tabular}[t]{l}
      \\
      \includegraphics{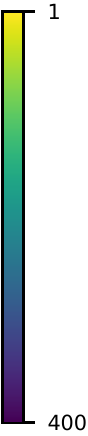} \\
      \\
      \includegraphics{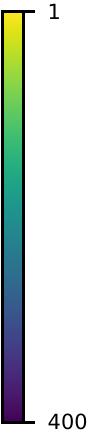} \\
      \\
      \includegraphics{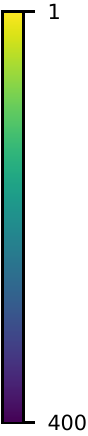} \\
    \end{tabular} &
    \begin{tabular}[t]{l}
      \multicolumn{1}{c}{\textbf{Std.-dev. ``Precision''}} \\
      \includegraphics{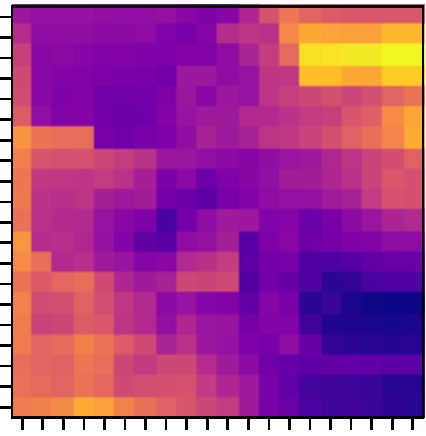} \\
      \multicolumn{1}{c}{\textbf{Std.-dev. ``Recall''}} \\
      \includegraphics{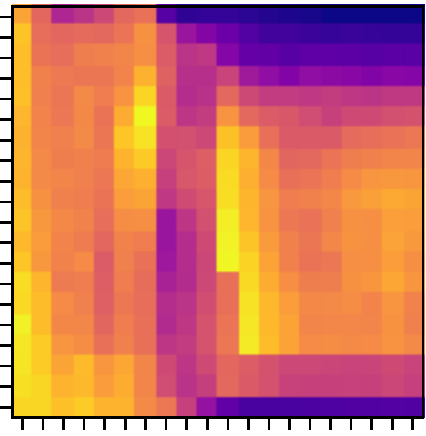} \\
      \multicolumn{1}{c}{\textbf{Std.-dev. ``F1''}} \\
      \includegraphics{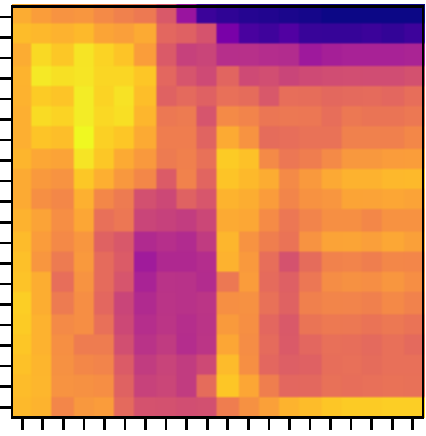} \\
      \includegraphics{img/m-estimate/x_axis_labels.pdf} \\
    \end{tabular} &
    \begin{tabular}[t]{l}
      \\
      \includegraphics{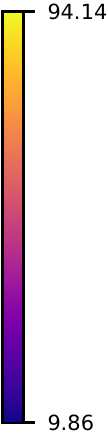} \\
      \\
      \includegraphics{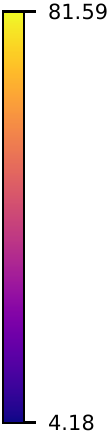} \\
      \\
      \includegraphics{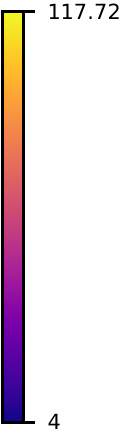} \\
      $m$ \\
    \end{tabular} \\
  \end{tabular}
\caption{Ranks and standard deviation of average ranks over all data sets according to micro-averaged precision, recall, and F1-measure. Best parameters for different data sets specified by red \textcolor{red}{+} signs.}
\label{fig_evaluation_avg_micro_measures}
\end{figure}

Some of the used data sets (\textsc{cal500}, \textsc{flags}, and \textsc{yeast}) contain very frequent labels for which the minority class $t_i = 0$. This is rather atypical in MLC and causes the unintuitive effect that the removal of individual rules results in a theory with greater recall and/or lower precision. To be able to compare different parameter settings across multiple data sets, we worked around this effect by altering affected data sets., i.e., inverting all labels for which $t_i = 0$.

\subsubsection{Predictive performance.}
\label{sec_predictive_performance}

In Figure~\ref{fig_evaluation_avg_hamm_subs} and \ref{fig_evaluation_avg_micro_measures} the average ranks of the tested configurations according to different performance measures are depicted. The rank of each of the 400 parameter settings was determined for each data set separately and then averaged over all data sets. The depicted standard deviations show that the optimal parameter settings for a respective measure may vary depending on the data set. However, for each measure there is an area in the parameter space where a good setting can be found with high certainty.

\begin{figure}[b!]
  \centering
  \begin{tabular}{lllll}
    \begin{tabular}[t]{l}
      \multicolumn{1}{c}{$\phi$} \\
      \includegraphics{img/y_axis_labels.pdf} \\
    \end{tabular}
    \begin{tabular}[t]{l}
      \multicolumn{1}{c}{\textbf{Avg. ranks ``F1''}} \\
      \includegraphics{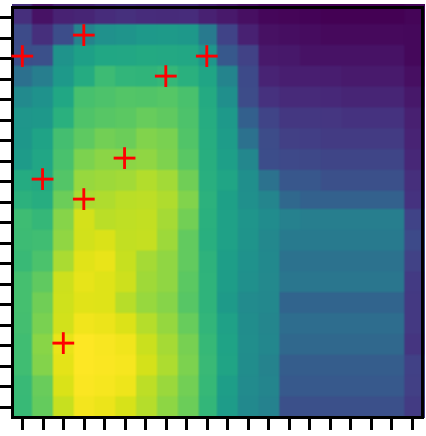} \\ 
      \includegraphics{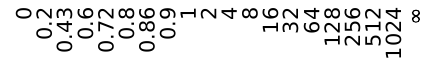} \\
    \end{tabular} &
    \begin{tabular}[t]{l}
      \\
      \includegraphics{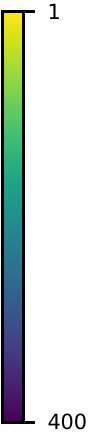} \\
    \end{tabular} &
    \begin{tabular}[t]{l}
      \multicolumn{1}{c}{\textbf{Std.-dev. ``F1''}} \\
      \includegraphics{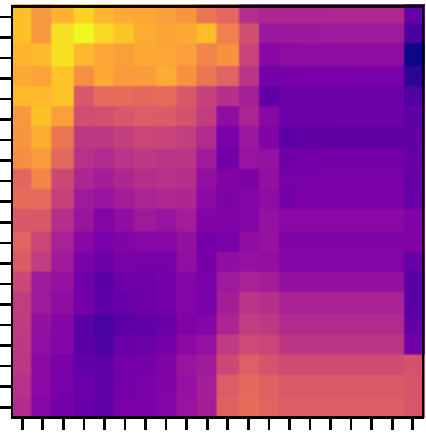} \\
      \includegraphics{img/f-measure/x_axis_labels.pdf} \\
    \end{tabular} &
    \begin{tabular}[t]{l}
      \\
      \includegraphics{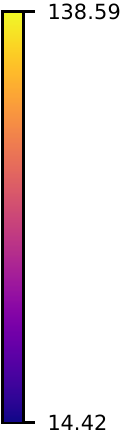} \\
      $\beta$ \\
    \end{tabular} \\
  \end{tabular}
\caption{Ranks and standard deviation of average ranks over all data sets according to micro-averaged F1-measure, when using the F-measure with varying $\beta$-parameters (horizontal axis) instead of the m-estimate for candidate selection. Best parameters for different data sets specified by red \textcolor{red}{+} signs.}
\label{fig_evaluation_avg_f1}
\end{figure}

As it can clearly be seen, precision and recall are competing measures. The first is maximized by choosing small values for $m$ and filtering extensively, the latter benefits from large values for $m$ and no filtering. Interestingly, setting $m = 0$, i.e., selecting candidates according to the precision metric, does not result in models with the highest overall precision. This is in accordance with Figure~\ref{fig_evaluation_avg_f1}, where the models with the highest F1 score do not result from using the F1-measure for candidate selection. Instead, optimizing the F1 score requires to choose small values for $m$ to trade off between consistency and coverage. The same applies to Hamming and subset accuracy, albeit both of these measure demand to put even more weight on consistency and filtering more extensively compared to F1.

\begin{figure}[p]
  \centering
  \begin{tabular}{llll}
    \begin{tabular}[t]{l}
      \multicolumn{1}{c}{$\phi$}\\
      \includegraphics{img/y_axis_labels.pdf} \\
      \\
      \includegraphics{img/y_axis_labels.pdf} \\
    \end{tabular}
    \begin{tabular}[t]{l}
      \multicolumn{1}{c}{\textbf{Avg. ranks ``Rules''}} \\
      \includegraphics{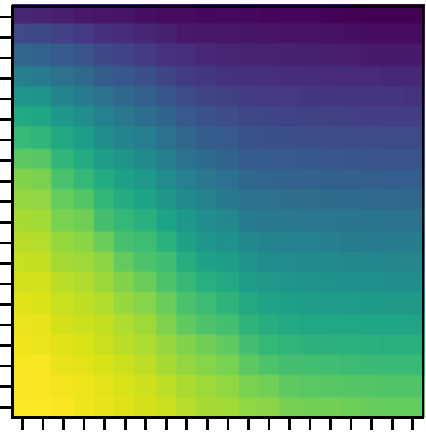} \\
      \multicolumn{1}{c}{\textbf{Avg. ranks ``Conditions''}} \\
      \includegraphics{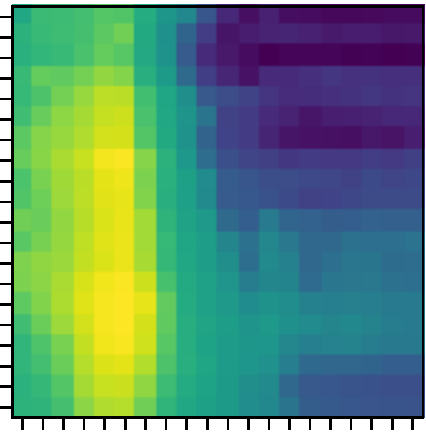} \\ 
      \includegraphics{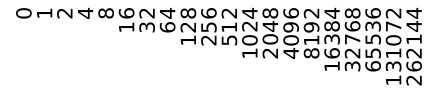} \\
    \end{tabular} &
    \begin{tabular}[t]{l}
      \\
      \includegraphics{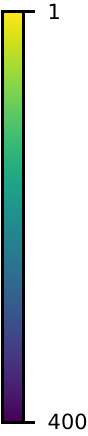} \\
      \\
      \includegraphics{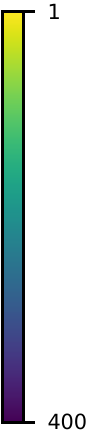} \\
    \end{tabular}
    \begin{tabular}[t]{l}
      \multicolumn{1}{c}{\textbf{Std.-dev. ``Rules''}} \\
      \includegraphics{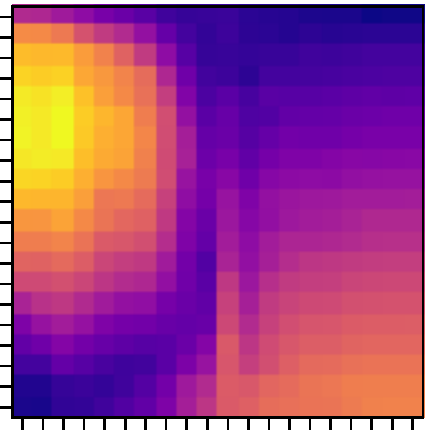} \\
      \multicolumn{1}{c}{\textbf{Std.-dev. ``Conditions''}} \\
      \includegraphics{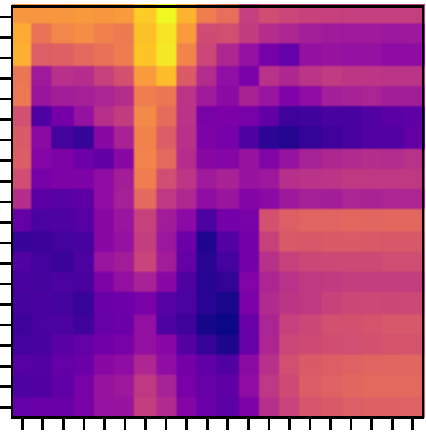} \\
      \includegraphics{figures/evaluation/x_axis_labels_m-estimate.pdf} \\
    \end{tabular} &
    \begin{tabular}[t]{l}
      \\
      \includegraphics{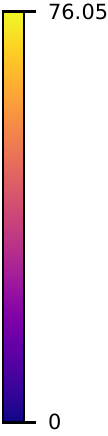} \\
      \\
      \includegraphics{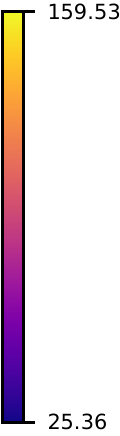} \\
      $m$ \\
    \end{tabular} \\
  \end{tabular}
\caption{Ranks and standard deviation of average ranks over all data sets regarding the number of rules and conditions. A smaller rank means more rules or conditions.}
\label{fig_evaluation_avg_number_rules_conditions}
\end{figure}

\begin{figure}[p]
  \centering
  \begin{tabularx}{\textwidth}{X r}
    \hline
    $m = 16, \phi = 0.3$ & Mi. Precision = 74.07\%, Mi. Recall = 78.26\% \\
  \end{tabularx}
  \begin{tabularx}{\textwidth}{r X}
    \hline
    \noalign{\smallskip}
    $Cough \leftarrow$ & \multicolumn{1}{p{10cm}}{\raggedright $\cond{cough} \wedge \cond{aldrich} \wedge \cond{opacity} \wedge \ncond{tachypnea} \wedge \ncond{streaky} \wedge \ncond{side} \wedge \ncond{distal} \wedge \ncond{diaphragm}$} \\
    $Cough \leftarrow$ & \multicolumn{1}{p{10cm}}{\raggedright $\cond{cough} \wedge \ncond{x-rays} \wedge \ncond{vomiting} \wedge \ncond{proximity} \wedge \ncond{hematuria} \wedge \ncond{focal}$} \\
    $Cough \leftarrow$ & \multicolumn{1}{p{10cm}}{\raggedright $\cond{cough} \wedge \ncond{group} \wedge \ncond{edema} \wedge \ncond{fever}$} \\
    $Cough \leftarrow$ & \multicolumn{1}{p{10cm}}{\raggedright $\cond{cough} \wedge \ncond{lobe} \wedge \ncond{breathing}$} \\
    $Cough \leftarrow$ & \multicolumn{1}{p{10cm}}{\raggedright $\cond{coughing}$} \\
  \end{tabularx}
  \begin{tabularx}{\textwidth}{X r}
    \hline
    $m = 262144, \phi = 1.0$ & Mi. Precision = 65.61\%, Mi. Recall = 89.57\% \\
  \end{tabularx}
  \begin{tabularx}{\textwidth}{r X}
    \hline
    \noalign{\smallskip}
    $Cough \leftarrow$ & \multicolumn{1}{p{10cm}}{\raggedright $\cond{cough} \wedge \ncond{ureteral} \wedge \ncond{stones} \wedge \ncond{contrast}$} \\
    $Cough \leftarrow$ & \multicolumn{1}{p{10cm}}{\raggedright $\cond{coughing}$} \\
    $Cough \leftarrow$ & \multicolumn{1}{p{10cm}}{\raggedright $\cond{code}$} \\
    $Cough \leftarrow$ & \multicolumn{1}{p{10cm}}{\raggedright $\cond{substance}$} \\
  \end{tabularx}
  \caption{Exemplary rule sets predicting the label \emph{786.2:Cough} of the data set \textsc{medical}, which contains textual radiology reports that were categorized into diseases.}
  \label{fig_example_rulesets}
\end{figure}

\subsubsection{Model characteristics.}
\label{sec_model_characteristics}

Besides the predictive performance, we are also interested in the characteristics of the theories. Figure~\ref{fig_evaluation_avg_number_rules_conditions} shows how the number of rules in a theory as well as the average number of conditions are affected by varying parameter settings. The number of rules independently declines when using greater values for the parameter $m$ and/or smaller values for $\phi$. resulting in less complex theories that can be comprehended by humans more easily. The average number of conditions is mostly affected by the parameter $m$.

Figure~\ref{fig_example_rulesets} provides an example of how different parameters affect the model characteristics. It shows the rules for predicting the same label as induced by two fundamentally different approaches. The first approach ($m = 16, \phi = 0.3$) reaches high scores according to the F1-measure, Hamming accuracy, and subset accuracy, whereas the second one ($m = 262144, \phi = 1.0$) results in high recall.

\subsection{Baseline comparison}
\label{sec_baseline_comparison}

Although the goal of this work is not to develop a method that generally outperforms existing rule learners, we want to ensure that we achieve competitive results. For this reason, we compared our method to JRip, Weka's re-implementation of Ripper \cite{cohen1995}, using the binary relevance method. By default, Ripper uses \emph{incremental reduced error pruning} (IREP) and post-processes the induced rule set. Although our approach could make use of such optimizations, this is out of the scope of this work. For a fair comparison, we also report the results of JRip without using IREP ($P = \textit{false}$) and/or with post-processing turned off ($O = 0$).

\begin{table}[b!]
  \centering
    \caption{Predictive performance of Ripper using IREP and post-processing ($R_3$), without using post-processing ($R_2$), and using neither IREP nor post-processing ($R_1$) compared to approaches trying to optimize micro-averaged F1 ($M_F$), Hamming accuracy ($M_H$), and subset accuracy ($M_S$).}
  \label{table_baselines}
  \resizebox{\textwidth}{!}{\setlength{\tabcolsep}{3pt}\begin{tabular}{ l|r r r r|r r r r|r r r r }
                      & \multicolumn{4}{c|}{\textbf{F1}} & \multicolumn{4}{c|}{\textbf{Hamming acc.}} & \multicolumn{4}{c}{\textbf{Subset acc.}} \\
                      & \multicolumn{1}{c}{$R_1$} & \multicolumn{1}{c}{$R_2$} & \multicolumn{1}{c}{$R_3$} & \multicolumn{1}{c|}{$M_F$} & \multicolumn{1}{c}{$R_1$} & \multicolumn{1}{c}{$R_2$} & \multicolumn{1}{c}{$R_3$} & \multicolumn{1}{c|}{$M_H$} & \multicolumn{1}{c}{$R_1$} & \multicolumn{1}{c}{$R_2$} & \multicolumn{1}{c}{$R_3$} & \multicolumn{1}{c}{$M_S$}  \\
    \hline
    \textsc{birds}    & \ttfamily 43.65 & \ttfamily 41.12 & \ttfamily\bfseries 46.01 & \ttfamily 45.33 & \ttfamily 94.39 & \ttfamily 94.48 & \ttfamily\bfseries 95.17 & \ttfamily 95.10 & \ttfamily 44.20 & \ttfamily 45.57 & \ttfamily\bfseries 51.48 & \ttfamily 48.85 \\ 
    \textsc{cal500}   & \ttfamily 33.63 & \ttfamily 33.18 & \ttfamily 33.76 & \ttfamily\bfseries 40.10 & \ttfamily 82.14 & \ttfamily 83.66 & \ttfamily 85.39 & \ttfamily\bfseries 86.02 & \ttfamily 0.00  & \ttfamily 0.00  & \ttfamily 0.00  & \ttfamily 0.00           \\
    \textsc{emotions} & \ttfamily 56.96 & \ttfamily 58.68 & \ttfamily 60.97 & \ttfamily\bfseries 65.20 & \ttfamily 75.12 & \ttfamily 75.38 & \ttfamily 77.21 & \ttfamily\bfseries 77.65 & \ttfamily 18.04 & \ttfamily 20.40 & \ttfamily\bfseries 23.60 & \ttfamily 22.42 \\
    \textsc{enron}    & \ttfamily 50.57 & \ttfamily 53.05 & \ttfamily\bfseries 55.33 & \ttfamily 51.07 & \ttfamily 94.35 & \ttfamily 94.70 & \ttfamily\bfseries 94.93 & \ttfamily 94.54 & \ttfamily 6.17  & \ttfamily 7.99  & \ttfamily\bfseries 9.16  & \ttfamily 7.81  \\
    \textsc{flags}    & \ttfamily 71.81 & \ttfamily 72.96 & \ttfamily\bfseries 74.85 & \ttfamily 72.83 & \ttfamily 73.02 & \ttfamily 74.08 & \ttfamily\bfseries 75.20 & \ttfamily 73.39 & \ttfamily 15.47 & \ttfamily 17.05 & \ttfamily\bfseries 21.00 & \ttfamily 9.82  \\
    \textsc{genbase}  & \ttfamily 98.83 & \ttfamily 98.68 & \ttfamily 98.68 & \ttfamily\bfseries 99.14 & \ttfamily 99.89 & \ttfamily 99.88 & \ttfamily 99.88 & \ttfamily\bfseries 99.92 & \ttfamily 97.28 & \ttfamily 96.83 & \ttfamily 96.83 & \ttfamily\bfseries 97.89 \\
    \textsc{medical}  & \ttfamily 81.40 & \ttfamily 83.67 & \ttfamily\bfseries 84.81 & \ttfamily 81.67 & \ttfamily 99.01 & \ttfamily 99.10 & \ttfamily\bfseries 99.15 & \ttfamily 98.98 & \ttfamily 66.74 & \ttfamily 69.91 & \ttfamily\bfseries 72.16 & \ttfamily 66.43 \\
    \textsc{scene}    & \ttfamily 63.97 & \ttfamily 63.25 & \ttfamily 64.55 & \ttfamily\bfseries 67.44 & \ttfamily 87.87 & \ttfamily 87.25 & \ttfamily 88.03 & \ttfamily\bfseries 88.93 & \ttfamily 46.61 & \ttfamily 44.54 & \ttfamily 46.24 & \ttfamily\bfseries 49.73 \\
    \textsc{yeast}    & \ttfamily 58.65 & \ttfamily 60.41 & \ttfamily 61.19 & \ttfamily\bfseries 64.25 & \ttfamily 78.50 & \ttfamily 78.29 & \ttfamily 78.77 & \ttfamily\bfseries 79.24 & \ttfamily 8.73  & \ttfamily 7.86  & \ttfamily 9.18  & \ttfamily\bfseries 11.75 \\
    \hline
    \textbf{Avg. rank} & \ttfamily 3.44 & \ttfamily 3.00 & \ttfamily 1.67 & \ttfamily 1.78 & \ttfamily 3.44 & \ttfamily 2.89 & \ttfamily 1.67 & \ttfamily 1.89 & \ttfamily 2.89 & \ttfamily 2.67 & \ttfamily 1.56 & \ttfamily 2.11 \\
  \end{tabular}}
\end{table}

Note that we do not consider the random forests from which we generate rules (cf.~Section~\ref{sec_rule_generation}) to be relevant baselines. This is, because random forests use voting for making a prediction, which is fundamentally different than rule learners that model a DNF. Also, we train random forests consisting of a very large number of trees with varying depths to generate diverse rules. In our experience, these random forests perform badly compared to commonly used configurations.

We tested three different configurations of our approach. The parameters $m$ and $\phi$ used by these approaches have been determined on a validation set by using nested 5-fold cross validation on the training data. For the approach $M_F$, the parameters have been chosen such that the F1-measure is maximized. The approaches $M_H$ and $M_S$ were tuned with respect to Hamming and subset accuracy, respectively.

According to Table~\ref{table_baselines}, our method is able to achieve reasonable predictive performances. With respect to the measure they try to optimize, our approaches generally rank before JRip with optimizations turned off ($R_1$), which is the competitor that is conceptually closest to our method. Although IREP definitely has a positive effect on the predictive performance, our approaches also tend to outperform JRip with IREP enabled, but without using post-processing ($R_2$). Despite the absence of advanced pruning and post-processing techniques, our approaches are even able to surpass the fully fledged variant of JRip ($R_1$) on some data sets. We consider these results as a clear indication that it is indispensable to be able to flexibly adapt the heuristic used by a rule learner --- which JRip is not capable of ---, if one aims at deliberately optimizing a specific multi-label performance measure.

\section{Related work}
\label{sec_related_work}

Several rule-based approaches to multi-label classification have been proposed in the literature. On the one hand, there are methods based on descriptive rule learning, such as association rule discovery \cite{thabtah2004, thabtah2006, li2008, lakkaraju2016}, genetic algorithms \cite{allamanis2013, cano2013}, or evolutionary classification systems \cite{arunadevi2011, avila2010}. On the other hand, there are algorithms that adopt the separate-and-conquer strategy used by many traditional rule learners for binary or multi-class classification, e.g. by Ripper \cite{cohen1995}, and transfer it to MLC \cite{mencia2016, rapp2018}. Whereas in descriptive rule learning one does usually not aim at discovering rules that minimize a certain (multi-label) loss, the latter approaches employ a heuristic-guided search for rules that optimize a given rule learning heuristic and hence could benefit from the results of this work.

Similar to our experiments, empirical studies aimed at discovering optimal rule learning heuristics have been published in the realm of single-label classification \cite{janssen2008, janssen2010}. Moreover, to investigate the properties of bipartition evaluation functions, ROC space isometrics have been proven to be a helpful tool \cite{flach2003, furnkranz2003}. They have successfully been used in the literature to study the effects of using different heuristics in separate-and-conquer algorithms \cite{furnkranz2005}, or for ranking and filtering rules \cite{furnkranz2004}.

\section{Conclusions}
\label{sec_conclusion}

In this work, we presented a first empirically study that thoroughly investigates the effects of using different rule learning heuristics for candidate selection and filtering in the context of multi-label classification. As commonly used multi-label measures, such as micro-averaged F1, Hamming accuracy, or subset accuracy, require to put more weight on the consistency of rules rather than on their coverage, models that perform well with respect to these measures are usually small and tend to contain specific rules. This is beneficial in terms of interpretability as less complex models are assumed to be easier to understand by humans.

As our main contribution, we emphasise the need to flexibly trade off the consistency and coverage of rules, e.g., by using parameterized heuristics like the m-estimate, depending on the multi-label measure that should be optimized by the model. Our study revealed that the choice of the heuristic is not straight-forward, because selecting rules that minimize a certain loss functions locally does not necessarily result in that loss being optimized globally. E.g., selecting rules according to the F1-measure does not result in the overall F1 score to be maximized. For optimal results, the trade-off between consistency and coverage should be fine-tuned depending on the data set at hand. However, our results indicate that, even across different domains, the optimal settings for maximizing a measure can often be found in the same region of the parameter space.

In this work, we restricted our study to DNFs, i.e., models that consist of non-conflicting rules that all predict the same outcome for an individual label. On the one hand, this restriction simplifies the implementation and comprehensibility of the learner, as no conflicts may arise at prediction time. On the other hand, we expect that including both, rules that model the presence as well as the absence of labels, could be beneficial in terms of robustness and could have similar, positive effects on the consistency of the models as the threshold selection used in this work. Furthermore, we leave the empirical analysis of macro-averaged performance measures for future work. 

\bibliography{bibliography}
\bibliographystyle{plain}

\end{document}